\documentclass[runningheads]{llncs}
\usepackage{graphicx}
\usepackage{comment}
\usepackage{amsmath,amssymb} 
\usepackage{color}

\usepackage[width=122mm,left=12mm,paperwidth=146mm,height=193mm,top=12mm,paperheight=217mm]{geometry}

\newcommand{\DF}{\emph{DeepFashion}~}
\newcommand{\StoS}{\emph{Street2Shop}~}
\newcommand{\ti}{\textit}

\begin{document}
\pagestyle{headings}
\mainmatter
\def\ECCVSubNumber{6313}  

\title{A Strong Baseline for Fashion Retrieval with Person Re-Identification models} 


\titlerunning{A Strong Baseline for Fashion Retrieval with Person Re-Identification models}
\author{Mikolaj Wieczorek\inst{1} \and
Andrzej Michalowski\inst{1} \and
Anna Wroblewska\inst{1,2}\orcidID{0000-0002-3407-7570} \and 
Jacek Dabrowski\inst{1}\orcidID{0000-0002-1581-2365}}
\authorrunning{M. Wieczorek et al.}
%
\institute{Synerise \and 
Faculty of Mathematics and Information Science, Warsaw University of Technology}

\maketitle

\begin{abstract}
Fashion retrieval is the challenging task of finding an exact match for fashion items contained within an image. Difficulties arise from the fine-grained nature of clothing items, very large intra-class and inter-class variance. Additionally, query and source images for the task usually come from different domains - street photos and catalogue photos respectively. Due to these differences, a significant gap in quality, lighting, contrast, background clutter and item presentation exists between domains. As a result, fashion retrieval is an active field of research both in academia and the industry.

Inspired by recent advancements in Person Re-Identification research, we adapt leading ReID models to be used in fashion retrieval tasks. We introduce a simple baseline model for fashion retrieval, significantly outperforming previous state-of-the-art results despite a much simpler architecture. We conduct in-depth experiments on \StoS and \DF datasets and validate our results. Finally, we propose a cross-domain (cross-dataset) evaluation method to test the robustness of fashion retrieval models.


\keywords{clothes retrieval, quadruplet loss, person re-identification, deep learning in fashion}
\end{abstract}

\section{Introduction}

Fashion image retrieval,\footnote{Image retrieval pertains to finding similar images as a whole, while in the case of fashion, the task is an instance retrieval task, as we want to find a match to a single item contained in the image. In this work we use those two terms interchangeably.} commonly associated with visual search, has received a lot of attention recently. This is an interesting challenge from both a commercial and academic perspective.
The task of fashion retrieval is to find an exact or very similar products from the vendor's catalogue (gallery) to the given query image. Creating a model that can find similarities between content of the images is essential for the two basic visual-related products for the fashion industry: recommendations and search. 

Visual recommendations (VR) - visual similarity needs to be found between a currently viewed product and other products in the database. Retrieval is done among the images from a single domain, which is a domain of catalogue photos.

Visual search (VS) - visual similarity between user taken/uploaded photo and the products' photos in the vendor's database. This case is a cross-domain retrieval as the photos from catalogue contains generally a single item and are taken by professionals using high-quality equipment, assuring proper background (usually solid white) and lighting conditions. On the other hand, photos taken by users are taken with a smartphone camera in uncontrolled lighting conditions and are of lower quality, noisy, with multiple persons and garments.

From a business perspective, especially for e-commerce retailers, these tools still have a lot of untapped potential. VS in particular can be used in various ways to enrich the customer experience and facilitate a more convenient search, since it may be easier and more natural for users to take a photo of a product and upload it directly via an app rather than to use textual search. 

During work on VS solutions, we found that the task of visual search for clothes is in many ways analogous to Person Re-Identification problem (ReID). Yet we have not encountered any work that examines models specific to Person ReID tasks in the context of fashion image retrieval tasks. Thus, we decided to modify and apply approaches from Person ReID field to fashion retrieval.

There are four main differences between the ReID and VS problems. Firstly, in ReID tasks data may be regarded as homogeneous, since the query and gallery images are both taken by CCTV cameras, therefore they are from the same domain. The cameras, resolution, view angles, and other factors may vary within the domain. In fashion visual search tasks images come from two domains. One domain contains store catalogue photos taken by professionals at the same studio settings. The second domain consists of photos taken by users, which may vary due to lighting conditions, the quality of the camera, angles and focal length. Additionally, the content of the image may be photographed differently. Catalogue photos are often well aligned, the background is usually solid white or monochromatic and the fashion item is fully visible. In contrast, user images often contain multiple objects, a cluttered background and the fashion item is often occluded or cropped.

Secondly, ReID problems may use additional temporal information to narrow down the search space, e.g. \cite{wang_spatial-temporal_2018}, which adds an information stream unavailable in the fashion domain. 

Moreover, fashion items, especially clothes, are highly deformable goods resulting in a high intra-class variance in appearance, further complicating appearance comparisons.

Finally, in ReID, the majority of images contain a whole person and can be relatively easily aligned. User images in fashion may differ both in the alignment, orientation and content. They may contain a whole person with multiple clothing items, upper or lower half of a person, clothes laying on the sofa etc. 

\begin{figure}
\begin{center}
\includegraphics[width=0.7\textwidth]{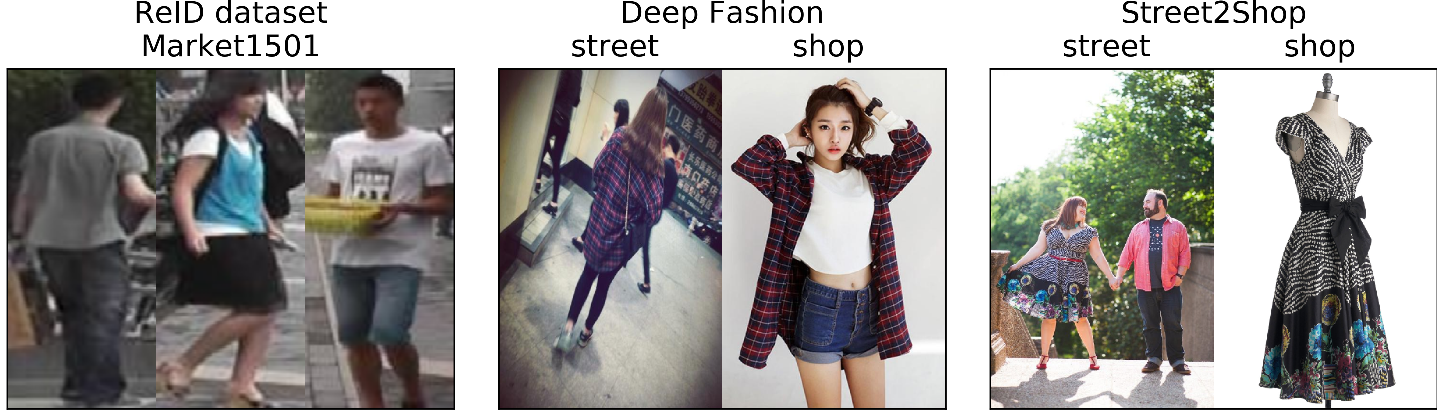}
\caption{Examples of images from ReID dataset, \DF and \StoS. The difference in alignment, background and quality can be seen between ReID and clothes retrieval datasets, as well as between street and shop photos within a single dataset.}
\label{fig:reid_vs_fashion}
\end{center}
\end{figure}

We also pinpointed four major similarities between ReID and clothes retrieval. 
In both domains, the core problem is an instance retrieval task that relies heavily on fine-grain details, which makes both tasks complex and challenging.
Secondly, both domains must implicitly consider a common subset of clothes \& materials deformations specific to human body poses and shapes.
Moreover, in both domains the models have to deal with occlusion, lighting, angle and orientation differences.
Finally, clothes are also an important factor in ReID task, especially when faces are not visible.

Due to the similarities and despite the differences of the two aforementioned tasks we decided to investigate how and which ReID solutions can be effectively applied to the fashion retrieval task.
Our main contributions are:
\begin{itemize}
    \item Recognition of deep analogies between Person ReID and Fashion Retrieval tasks.
    \item A review of ReID models and their applicability to Fashion Retrieval tasks. 
    \item Adjustments of ReID models for their application to the fashion domain.
    \item Thorough evaluation of the adapted ReID models on two commonly used fashion datasets.
    \item A simple baseline model for fashion retrieval significantly outperforming state-of-the-art results for \DF
    and \StoS datasets.
    \item A cross-domain (cross-dataset) evaluation method to test robustness of the solutions.
\end{itemize}

In the following sections we provide a review of Person Re-Identification models and solutions used in Fashion Retrieval domain. We show the baselines for our comparisons (Section~\ref{sec:related_work}). Then, we provide an argument for our selection of applicable ReID models (Section~\ref{sec:our_approach}). Subsequently, we describe our results, compare them with previous state-of-the-art approaches and provide a discussion (Section~\ref{sec:experiments}) of our findings. We conclude with our best achievements (Section~\ref{sec:conclusions}). In supplement material we provide statistics of datasets used for training and evaluation, as well as additional samples from model outputs.  

\section{Related Work}\label{sec:related_work}
This section focuses mainly on the problems of extracting features from images and the retrieval process itself. 
Instance retrieval is basically a two-stage process: (1) encoding image/item into a n-dimensional embedding vector; (2) searching for the most similar product embedding to the query in the n-dimensional space under a chosen distance metric.

\subsection{Feature Extraction}
The process of extracting features from an image may be defined as passing the pixel matrix through a function to obtain an embedding vector that encompasses all necessary information about the image in the encoded form. It is the first and most indispensable step towards image retrieval. Most research is devoted to improving the performance of this stage. 

Over the years, the functions used for encoding image content have evolved. At first, hand-crafted features describing global image characteristics (e.g. histogram of oriented gradients - HOG) or local dependencies like SIFT (scale-invariant feature transform) were used. These methods did work, however subsequently convolutional neural networks (CNNs) surpassed these methods in computer vision tasks and dominated the field.


One of the first works using CNN for an image retrieval task was \cite{razavian_cnn_2014} that also managed to surpass the stat-of-the-art of approaches based on SIFT, HOG and other hand-crafted features. In \cite{gordo_end--end_2017} authors used CNNs extracted features and trained a model using a triplet loss function, which was widespread by \cite{schroff_facenet:_2015} for face recognition task. 
By 'CNN extracted features' we mean the final output of either one or many convolutional layers that are arranged in various combinations (e.g. pyramid, sequence of layers). All of these combinations work as various learned filters for an image matrix.

\subsection{Image Retrieval}\label{subsec:related_work:image_retrieval}

Image retrieval may be defined as a task of matching images based on their feature vectors/embeddings under a chosen distance metric.
There are three main approaches when dealing with image retrieval using deep learning embeddings: direct representation, classification loss training and deep metric learning. We explain them in the following paragraphs.

\paragraph{Direct representation}
This is the simplest approach as it does not involve any training on the target dataset. In most cases, an ImageNet pre-trained neural network is used. To obtain the feature vector, an image is fed into the network and the output from one or more layers is treated as its embedding. To find the most similar images to the query, a cosine or other similarity metric is applied to the set of embeddings. Such an approach was used in \cite{kiapour_where_2015}.


\paragraph{Classification loss training}
Another approach is to again take a pre-trained CNN and train it on the target dataset using a classification loss function. The difference to the direct representation is that the network is further trained (fine-tuned) on a target dataset. Thus, the embeddings should be optimized in a way to allow correct classification and in turn indirectly improve the retrieval quality.  

This approach was also present in previously mentioned reference \cite{kiapour_where_2015}, where the authors used pre-trained AlexNet to learn the similarity between street and shop photos using cross-entropy loss over pair of images, classifying them as either matching or non-matching. 

\paragraph{Deep metric learning}
Image retrieval formulated as a classification problem is not an optimal approach, as the classification is done mainly in the last fully connected (FC) layer of the neural network (NN), which is not used during the inference stage. Moreover, the gradient flow does not explicitly optimize the feature embeddings to be similar/dissimilar under the distance metric used for retrieval.

Deep metric learning treats the image retrieval problem somewhat as a clustering or ranking problem and optimizes the mutual arrangement of the embeddings in space. One of the most widely used approaches that pulls the same class closely and pushes away other class embeddings is using a triplet loss for training neural network. The triplet loss and its' modifications are described in our supplement in more details.

\subsection{Fashion Image Understanding}
Fashion image understanding has been an area of research for a while. The papers that introduced two modern large public datasets \cite{yamaguchi_paper_2013}, \cite{jagadeesh_large_2014}, gave the start for a rapid development of machine learning applications in the fashion field. Over the years numerous fashion datasets were released by various authors: \cite{kiapour_where_2015}, \cite{zheng_modanet_2018}, \cite{zou_fashionai:_nodate}, \cite{huang_cross-domain_2015}, \cite{liu_deepfashion:_2016}, \cite{ge_deepfashion2:_2019}, \cite{guo_imaterialist_2019} accelerating the research and allowing a wider range of tasks to be tackled.

Clothes retrieval seems to be the most popular task \cite{huang_cross-domain_2015}, \cite{ge_deepfashion2:_2019}, \cite{zheng_modanet_2018}, \cite{yang_clothing_2014}, \cite{kucer_detect-then-retrieve_2019}.
However, it is often combined with another task to attain a synergy effect, either with landmark detection (detecting keypoints of clothes) \cite{leal-taixe_deep_2019}, attribute prediction \cite{liao_interpretable_2018}, \cite{leal-taixe_deep_2019}, or object detection \cite{huang_cross-domain_2015}, \cite{ge_deepfashion2:_2019}, \cite{kucer_detect-then-retrieve_2019}, \cite{kiapour_where_2015}.


\subsection{Re-identification}
The problem of Person Re-Identification, being an instance retrieval task, has undergone a similar evolution as the image retrieval domain, starting from using hand-crafted features \cite{kostinger_large_2012} and matching their embeddings under chosen distance metric. Currently CNN extracted features are the dominant approach. 

Similarly to Section \ref{subsec:related_work:image_retrieval} a variety of loss functions are used in ReID problems: classification loss \cite{zheng_mars_2016}, verification loss \cite{geng_deep_2016} and triplet loss \cite{liu_multi-scale_2016}. Triplet Loss performs best \cite{hermans_defense_2017} in most cases by a significant margin.

Due to the fact that the ReID problem focuses on images of people from CCTV cameras, who are mostly in a standing position, some works exploited this fact by adding this information during training. In \cite{zhang_alignedreid_2018} the authors used horizontal pooling of image stripes, while \cite{quan_auto-reid_2019} modified a network architecture search algorithm specifically for ReID task, that also integrated the human body in an upright position during training. \cite{su_pose-driven_2017}, \cite{sarfraz_pose-sensitive_2018} proposed using human/skeleton pose information along with global and local features for better alignment of features, thus, increasing retrieval performance. 

Moreover, Person Re-Identification problem settings allow to implicitly or explicitly use spatio-temporal information, as time constraint allows to eliminate a portion of irrelevant images during retrieval stage \cite{cho_joint_2019}, \cite{huang_camera_2016}. Currently spatio-temporal approach \cite{wang_spatial-temporal_2018} tops the leaderboard in Person ReID task on \textit{Market1501} - common dataset for Person Re-Identification task.


\section{Our Approach}\label{sec:our_approach}
The aim of our work was to investigate if and how ReID models can be successfully applied to a fashion retrieval task. Based on the performance of various models in Person Re-Identification task\footnote{\label{footnote:paperswithcode} \url{https://paperswithcode.com/task/person-re-identification}} and their suitability for the fashion retrieval problem, we selected the most appropriate models. The motivation behind our final choice was threefold:
\begin{enumerate}
    \item The chosen models should exhibit top performance on ReID datasets.
    \item The chosen models should cover different approaches with regards to the network architecture, training regime etc.
    \item The code had to be publicly available.
\end{enumerate}

\subsection{ReID Models}
\label{our_approach:reid_model_choice}
Two models were chosen based on the above criteria.

First, \cite{luo_strong_2019} presents an approach that combines the most efficient training approaches and sets a strong baseline for other ReID researchers, hence its name: \textit{ReID Strong Baseline} (RST). Through a thorough evaluation it showed that by combining simple architecture, global features and using training tricks (warm-up learning rate, random erasing augmentation, label smoothing, last stride, BNNeck, center loss; for more details see \cite{luo_strong_2019}) one can surpass state-of-the-art performance in the ReID domain.  


The second choice -\ti{OSNet-AIN} \cite{zhou_learning_generalisable_2019} uses features extracted from different scales, which seems to be a well-suited approach in a fine-grained instance retrieval. The authors devised a flexible fusion mechanism that allows the network to focus on local or global features that improves performance. This is a different approach from the first one, as both global and local features are used and the whole neural architecture was purpose-built for ReID task.

It is worth mentioning that some top scoring papers from website \emph{Paperswithcode} were not chosen as they either required temporal data \cite{wang_spatial-temporal_2018} or made an implicit assumptions of spatial alignment of objects in images \cite{quan_auto-reid_2019}. Neither approach was suitable to the clothes retrieval task and the datasets we used for evaluation.


\subsection{Settings for Our Experiments}
\label{sec:setting-exper}
To fully explore ReID models and their application to the fashion retrieval task we conducted a number of experiments with the two selected approaches. We explored several aspects with potential to influence the performance in the fashion domain:
\begin{itemize}
    \item backbone architecture (a feature extractor part of a model; see Figure \ref{fig:rst_diagram})
    \item triplet and quadruplet loss functions in RST
    \item size of input images
    \item re-ranking (post-processing methods to improve accuracy of retrieval)
    \item cross-domain evaluation (training on one dataset and testing on another)
\end{itemize}

Using the selected ReID models we dealt with three loss functions. As for the \ti{OSNet-AIN}, we followed the training regime proposed by its authors and we trained the model using only a classification loss.
For the RST model we used a loss function that consisted of three elements: a classification loss, a triplet loss and a center loss. Additionally, we tested if by replacing the triplet loss with a quadruplet loss one could improve the performance.\footnote{Triplet and quadruplet loss functions and their modifications are described in our supplementary material in more detail.}

\subsection{SOTA Models for Clothes Retrieval vs RST Model}

\paragraph{A detect-then-retrieve model for multi-domain fashion item retrieval} \cite{kucer_detect-then-retrieve_2019} showed SOTA performance for \StoS dataset. 
The model consists of Object Detector (Mask-RCNN), which is used to crop clothing items from shop images and a Retrieval Model that is built upon Resnet-50 backbone, but incorporates RMAC pipeline \cite{gordo_end--end_2017}. The RMAC is a global representation of images that uses CNN as a local feature extractor. Local features are extracted and max-pooled from multiple multi-scale overlapping regions covering whole image, creating  a single feature vector for each region. These region features are aggregated, normalized, transformed and then normalized again. 
Moreover, in \cite{kucer_detect-then-retrieve_2019} an ensemble model was also created that consisted of two single models, one trained using triplet loss and other one using AP loss. 
Input image sizes used in \cite{kucer_detect-then-retrieve_2019} were also much larger than used in our RST model as the images were resized to 800 pixels, which caused problem in processing them thought the network, since they were forced to process either a single image (AP loss) or a single triplet (triplet loss) at a time.
For more details see \cite{kucer_detect-then-retrieve_2019}.

\paragraph{Fashion Retrieval via Graph Reasoning Networks on a Similarity Pyramid} \cite{kuang_fashion_2019} achieved SOTA performance on \DF dataset. They based their solution on graphs and graph convolution. The architecture consists of three parts.
First is a CNN network, which extracts multi-scale (pyramid) features from numerous, overlapping windows that cover the whole image. What is important two images of the same item are fed into the network – street and shop images and they are processed together as a pair.
Second part is the  Similarity Computation, which computes region similarity between street and shop images for all possible local feature combinations at the same pyramid scale.
Next, a graph is built. Computed region similarities are the nodes, the relations between region similarities are the edges. Scalar edges weights are computed automatically based on the node, incoming and outgoing edges. 
Finally, the convolution operations are applied to the graph and classification loss is used for training. The aim of the network is to classify the pair of images as depicting the same clothes or not. Input size of images was 224x224, which is slightly smaller than used in our best performing RST model.
For more details see \cite{kuang_fashion_2019}.

\begin{figure}
\begin{center}
\includegraphics[width=0.9\textwidth]{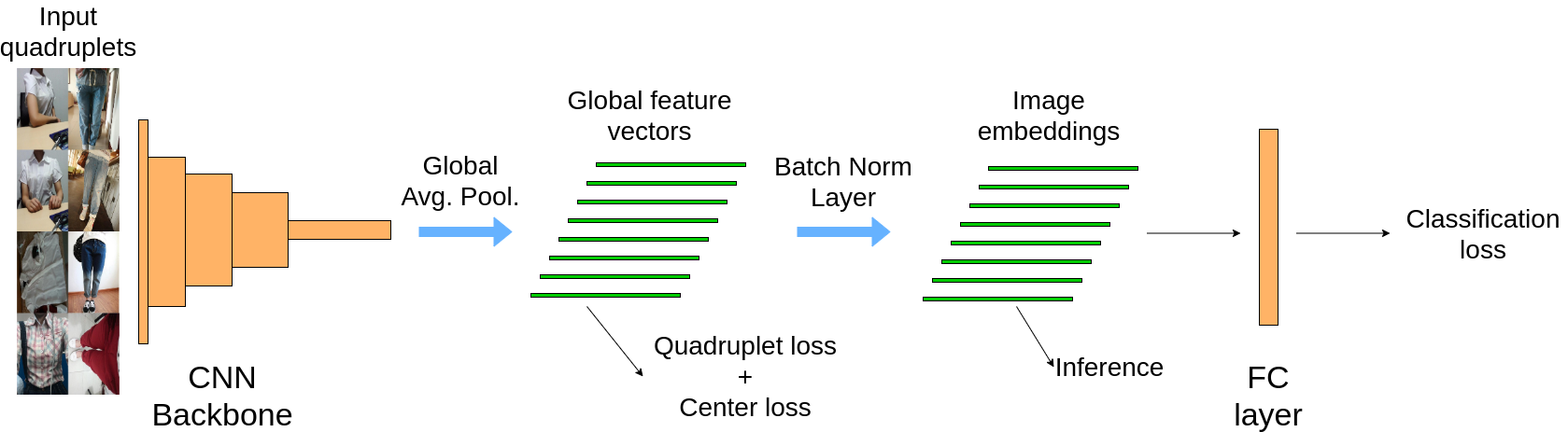}
\caption{RST model architecture}.
\label{fig:rst_diagram}
\end{center}
\end{figure}

\paragraph{RST model}
Compared to the two models described above, the RST model can be characterised by its simplicity. Even though, the architecture is much simpler, the performance exceeds significantly current SOTA.
The RST model consists of 3 parts. First, CNN backbone extracts features from images and global average pooling is applied to create global feature vectors. They are used for computing quadruplet and center loss. Next, the global feature vectors are normalized and we call them \textit{Images embeddings} (see Figure \ref{fig:rst_diagram}). \textit{Images embeddings} are used during training as input to a fully connected layer, while during inference stage (retrieval) they are used to compute similarity distance.
In Figure \ref{fig:rst_diagram} we presents the RST model, which shows the pipeline and all parts of the architecture. It can be deemed as strikingly simple, yet it substantially exceeds current SOTA results. 

\section{Experiments}
\label{sec:experiments}
In this section we describe our methodology and details of experiments we ran. We conducted evaluation of our approach and its settings on \StoS \\
and \DF datasets.\footnote{These datasets, used metrics and many visualizations are presented very precisely in our supplement.}

For both \StoS and \DF we cropped the clothing items of interest from the whole images, as both datasets contain ground-truth bounding boxes for street photos. Additionally \DF contains bounding boxes for shop images, so we cropped them as well. \StoS does not provide bounding boxes for shop images, therefore we used the original, un-cropped shop images for the gallery set. 
These settings for street images are in line with those used in \cite{kucer_detect-then-retrieve_2019} and \cite{kuang_fashion_2019}, which we consider as the current SOTA for \StoS and \DF respectively.

\subsection{Training}\label{subsec:training}

To adapt the RST model to fashion datasets we introduced some changes to the code, which were mostly required due to significant difference in size between ReID and clothes retrieval datasets. The size of \DF and \StoS \\ 
caused RAM and video RAM overflow, which was solved by using more memory-efficient methods. Moreover, we added some new functionalities and scripts. Our improvements are described in detail in the supplement.

Unfortunately, evaluation with re-ranking and without category constraint for \StoS was impossible despite improving the code efficiency, therefore we decided to conservatively estimate the results. The estimated results are marked with asterisk (*). For more details see the supplement.

\subsection{ReID Models Comparison}

In the first experiment we trained both ReID models to compare their performance on \DF and \StoS datasets. For training we used 256x128 input images. Table \ref{tab:models_comparision} contains the results of the models and current state-of-the-art approach. It can be seen that the RST model surpasses the current SOTA model and the \ti{OSNet-AIN} performs worse than the other two approaches. The reason for poor performance of \ti{OSNet} may be the fact that it is a lightweight network and built very specifically for ReID tasks.

Due to the large performance gap between RST and \ti{OSNet} models, we decided to conduct further experiments using only the more promising RST model.

\begin{table}[]
\begin{center}
\caption{Comparison of performance of models on \StoS and \DF data. Best performance across models on a given dataset is in bold}
\label{tab:models_comparision}
\begin{tabular}{l|cccc|ccccc}
                           & \multicolumn{4}{c|}{\StoS} & \multicolumn{5}{c}{\DF}       \\
Model                           & mAP   & Acc@1  & Acc@10 & Acc@20 & mAP  & Acc@1 & Acc@10 & Acc@20 & Acc@50 \\ \hline
RST              & \textbf{37.2}  & \textbf{42.3}   & \textbf{61.1}   & \textbf{66.5}   & \textbf{35}   & \textbf{30.8}  & \textbf{62.3}   & \textbf{69.4}   & \textbf{78}     \\
OSNet-AIN                  & 18.9  & 25.3   & 40.8   & 45.4   & 20.1 & 17.5  & 40.2   & 46.9   & 53.8   \\ \hline
Current SOTA               & 29.7  & 34.4   & -      & 60.4   & -    & 27.5  & -      & 65.3   & 75    
\end{tabular}
\end{center}
\end{table}

\subsection{Backbone Influence}\label{subsec:backbone}
In this section we inspect the influence of the used backbone on the results. The results of our experiments are shown in Table \ref{tab:backbone_influence}. All runs during this experiment were performed on input images of size 256x128, using an RST approach with all tricks enabled and using quadruplet loss. All models were trained for 120 epochs on 1 GPU, with base learning rate 0.0001 and batch size 128. \ti{SeResNext-50} \cite{hu_squeeze-and-excitation_2019}, \ti{EfficientNet-b2} and \ti{EfficientNet-b4} \cite{efficient_net_2019} were trained on 2 GPUs using ModelParallel mode due to their size. 

Our findings are in line with backbone performance presented by the authors in \cite{luo_strong_2019}, i.e. \ti{ResNet50-IBN-A} \cite{pan2018IBN-Net} is the best performing backbone. Regarding the results, we infer that such a large advantage in performance for \ti{Resnet50-IBNs}  may be caused by instance normalization, which reduces the impact of variations in contrast and lighting between street and shop photos. 
\begin{table}[]
\begin{center}
\caption{Results achieved by different backbones on \StoS and \DF datasets compared to the current SOTA. }
\label{tab:backbone_influence}
\begin{tabular}{l|cccc|ccccc}
                & \multicolumn{4}{c|}{\StoS} & \multicolumn{5}{c}{\DF} \\
Backbone        & mAP           & Acc@1         & Acc@10        & Acc@20 & mAP           & Acc@1         & Acc@10        & Acc@20        & Acc@50      \\ \hline
ResNet-50                & 32.0          & 36.6          & 55.3          & 60.6          & 32.4        & 28.1          & 58.3          & 65.5          & 74.2        \\
SeResNet-50              & 30.5          & 34.6          & 53.1          & 58.7          & 31.3        & 27.0          & 57.8          & 65.4          & 74.4        \\
SeResNeXt-50             & 31.9          & 36.9          & 54.5          & 59.7          & 32.2        & 27.8          & 58.5          & 66.0          & 74.6        \\
ResNet50-IBN-A           & \textbf{37.2} & \textbf{42,3} & \textbf{61.1} & \textbf{66.5} & \textbf{35.0} & \textbf{30.8} & 62.3          & \textbf{69.4} & \textbf{78.0} \\
ResNet50-IBN-B           & 36.9          & 41.9          & 60.6          & 65.1          & 32.2        & 28.1          & 58.4          & 65.8          & 74.7        \\
EfficientNet-b1          & 28.8          & 35.1          & 52.1          & 56.7          & 28.5        & 23.4          & 49.8          & 56.8          & 66.1        \\
EfficientNet-b2          & 29.4          & 34.0          & 50.8          & 56.2          & 24.1        & 20.4          & 45.9          & 53.2          & 62.6        \\
EfficientNet-b4          & 31.8          & 38.2          & 55.6          & 60.2          & 26.8        & 23.1          & 59.1          & 57.4          & 66.7        \\ \hline
Current SOTA             & 29.7          & 34.4          & -             & 60.4          & -           & 25.7          & \textbf{64.4} & -             & 75.0         
\end{tabular}
\end{center}
\end{table}

\subsection{Influence of Loss Functions}



The results for the RST model using triplet and quadruplet loss are shown in Table \ref{tab:loss_func_influence}. It can be seen that the quadruplet loss in our test settings performed marginally better than the triplet loss, yet it brings an improvement at almost no cost.

\begin{table}[]
\begin{center}
\caption{Our results on \StoS and \DF datasets achieved with triplet and quadruplet loss functions}
\label{tab:loss_func_influence}
\begin{tabular}{l|cccc|cccc}
              & \multicolumn{4}{c}{\StoS}                               & \multicolumn{4}{|c}{\DF}                           \\
Loss function & mAP           & Acc@1         & Acc@10        & Acc@20        & mAP         & Acc@1         & Acc@10        & Acc@20        \\ \hline
Quadruplet    & \textbf{37.2} & \textbf{42.3} & \textbf{61.1} & \textbf{66.5} & \textbf{35} & \textbf{30.8} & 62.3          & 69.4          \\
Triplet       & 37.1          & 41.8          & 60.4          & 65.7          & 34.8        & 30.5          & \textbf{62.4} & \textbf{69.5}
\end{tabular}
\end{center}
\end{table}



\subsection{Influence of Input Image Size}\label{subsec:input_size}

Even though the results achieved with 256x128 input images had already surpassed the current SOTA performance for many backbones (See Table \ref{tab:backbone_influence}), we decided to test if larger images would result in even higher performance. Outcomes of our experiments are presented in Table \ref{tab:input_sizes}. It can be seen that using larger images allows to further boost performance. In our settings, we achieved best results for input images of size 320x320. Using larger images (480x480) did not bring any advantage. Additional plots are available in the supplement.


\begin{table}[]
\begin{center}
\caption{Comparison of performance of clothes retrieval with different input image sizes. All experiments were performed using \ti{Resnet50-IBN-A} RST model with all tricks enabled, quadruplet loss function and without re-ranking}
\label{tab:input_sizes}
\begin{tabular}{l|cccc|ccccc}
             & \multicolumn{4}{c|}{\StoS}                              & \multicolumn{5}{c}{\DF}                                           \\
Input size   & mAP           & Acc@1         & Acc@10        & Acc@20        & mAP         & Acc@1         & Acc@10        & Acc@20        & Acc@50          \\ \hline
256x128      & 38.6          & 44.5          & 62.5          & 67.2          & 35.0        & 30.8          & 62.3          & 69.4          & 78.0            \\
224x224      & 42.4          & 49.2          & 66.2          & 70.9          & 40.5        & 35.8          & 68.5          & 75.1          & 82.4            \\
320x320      & \textbf{46.8} & \textbf{53.7} & \textbf{69.8} & \textbf{73.6} & \textbf{43.0} & \textbf{37.8} & \textbf{71.1} & \textbf{77.2} & \textbf{84.1} \\
480x480      & 46.6          & 53.5          & 69.0          & 72.9          & 42.4        & 37.3          & 69.4          & 75.4          & 82.2             \\ \hline
Current SOTA & 29.7          & 34.4          & -             & 60.4          & -           & 27.5          & -             & 65.3          & 76.0           
\end{tabular}
\end{center}
\end{table}

\begin{figure}
\begin{center}
\includegraphics[width=0.9\textwidth]{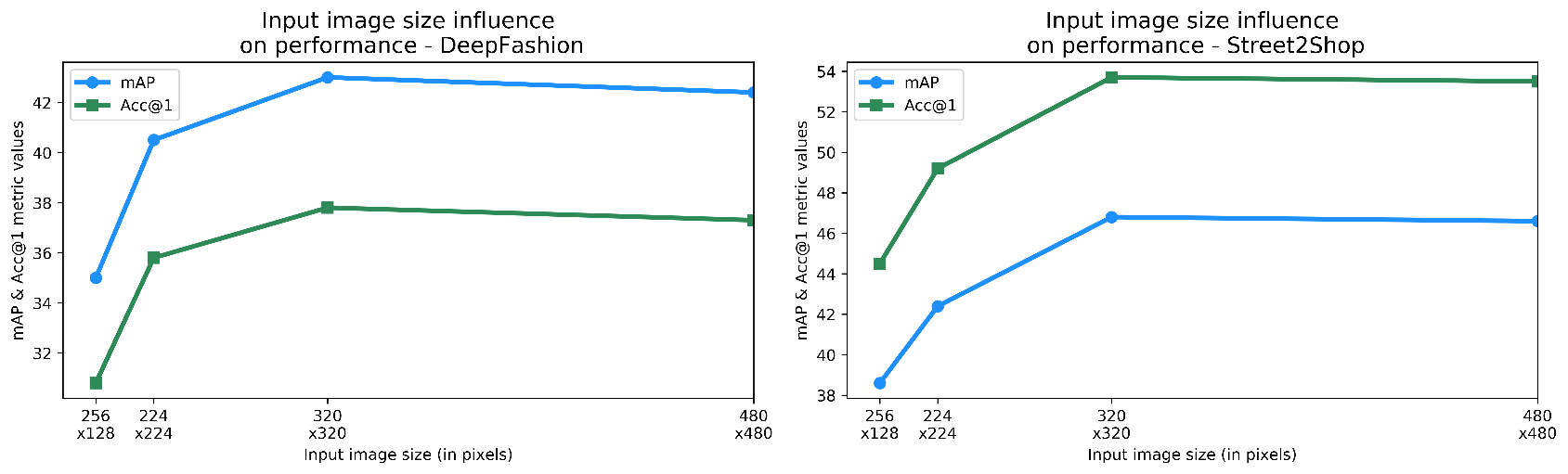}
\caption{Two plots presenting influence of input image size on mAP and Acc@1 metric for \DF and \StoS datasets respectively.}
\label{fig:metircs_plot}
\end{center}
\end{figure}

\subsection{Influence of Re-ranking}

One of the methods boosting performance tested in \cite{luo_strong_2019} was re-ranking during test stage. Re-ranking is a post-processing method applied to raw retrieval results in order to improve the accuracy of the results. In our experiments we used k-reciprocal encoding proposed by \cite{zhong_re-ranking_2017} similarly to \cite{luo_strong_2019}. 

Table \ref{tab:re_ranking} presents the comparison of RST models with and without using re-ranking. It can be seen that the results with re-ranking post-processing improved significantly for both datasets. The improvement is greatest for mAP and Acc@1, thus suggesting that re-ranking 'pushes' the positive samples closer to the start of the results list. Such behaviour seems desirable for real-world applications where a visual search user will obtain more relevant items at the top
\begin{table}[]
\begin{center}
\caption{The comparison of performance for the model without and with re-ranking. For results marked with * see Section \ref{subsec:training}}.
\label{tab:re_ranking}
\begin{tabular}{l|cccc|cccc}
              & \multicolumn{4}{c|}{Street2Shop}                                 & \multicolumn{4}{c}{DeepFashion}                          \\
              & mAP            & Acc@1          & Acc@10         & Acc@20        & mAP           & Acc@1       & Acc@10        & Acc@20      \\ \hline
No re-ranking & 46.8           & 53.7           & 69.8           & \textbf{73.6} & 43.0          & 37.8        & 71.1          & 77.2        \\
Re-ranking    & \textbf{54.8*} & \textbf{57.1*} & \textbf{69.9*} & 72.9*         & \textbf{47.3} & \textbf{40.0} & \textbf{73.5} & \textbf{79.0}
\end{tabular}
\end{center}
\end{table}

\subsection{Comparison to the State-of-the-art}
In Table \ref{tab:sota_comparison_s2s} we compare our results to the state-of-the-art on \StoS dataset \cite{kucer_detect-then-retrieve_2019}. 
We list the best performing ensemble model from the aforementioned paper, as it achieved better results than their best single model. In comparison, we show the results of our best performing single model, which was trained using an RST approach, \ti{Resnet50-IBN-A} backbone using quadruplet loss during training. It can be seen that generally our approach performs better by a large margin despite using only 320x320 input image size compared to 800x800 used in \cite{kucer_detect-then-retrieve_2019}.

In all categories and in overall performance our model clearly outperforms current state-of-the-art by a large margin. In two categories our model performs marginally worse, but it may be attributed to the fact that we use much smaller images. The categories with inferior performance consist of small items such as belts or eye-wear, where fine-grained details are more important and a high image resolution is beneficial. In our case cropped items are of very small dimensions, which may limit performance.

It is important to note that the mAP, Acc@1, Acc@20 reported by \cite{kucer_detect-then-retrieve_2019} as overall performance are just average values across all categories for each metric. There, the retrieval stage was limited to products only from a specific category, thus limiting the choice of the model and making retrieval less challenging. We also report the performance in these settings for our model and for \cite{kucer_detect-then-retrieve_2019} in the \textit{Average over categories} row of Table~\ref{tab:sota_comparison_s2s} to allow fair comparison.
Additionally, we propose an unconstrained evaluation, where we conduct retrieval from all gallery images, with no restrictions. Our results are in the  \textit{Unconstrained retrieval} row of Table~\ref{tab:sota_comparison_s2s}.

The large gap between  \textit{Average over categories} and  \textit{Unconstrained retrieval} derives from the fact that the well performing categories such as \textit{skirts} and \textit{dresses} are the most numerous ones, therefore they weigh more in the final results than categories that have few queries, such as  \textit{belts}. Hence, unconstrained retrieval is closer to a weighted average than the simple average named by us as \textit{Average over categories}, where each category has equal weight when calculating the final results. 


\begin{table}[]
\begin{center}
\caption{Comparison of performance on \StoS dataset using mAP, Acc@1, and Acc@20 metrics. Best performance for each category is presented in bold per metric. For results marked with * see Section \ref{subsec:training}}
\label{tab:sota_comparison_s2s}
\begin{tabular}{l|ccc|ccc|ccc}
                                                                   & \multicolumn{3}{c|}{Current SOTA \cite{kucer_detect-then-retrieve_2019}}   & \multicolumn{3}{c|}{Our Model}                 & \multicolumn{3}{c}{Our Model Re-ranking}       \\ \hline
Category                                                           & mAP  & Acc@1          & Acc@20        & mAP           & Acc@1         & Acc@20        & mAP            & Acc@1          & Acc@20        \\ \hline
bags                                                               & 23.4 & 36             & 62.6          & 32.2          & 44.2          & \textbf{74.6} & \textbf{39}    & \textbf{45.7}  & 69.6          \\
belts                                                              & 9.4  & 9.5            & \textbf{42.9} & \textbf{11.3} & \textbf{12.2} & 46.3          & 10.5           & 7.3            & 29.3          \\
dresses                                                            & 49.5 & 56.4           & 72.0          & 65.8          & 73.7          & \textbf{85.9} & \textbf{75.3}  & \textbf{76.7}  & 85.8          \\
eyewear                                                            & 26.7 & 36.2           & 91.4          & 27            & 31.4          & 76.5          & \textbf{27.5}  & \textbf{37.3}  & \textbf{80.4} \\
footwear                                                           & 11.0 & 14.8           & 34.2          & 34.2          & 37.9          & \textbf{65.4} & \textbf{44.3}  & \textbf{43.3}  & \textbf{65.4} \\
hats                                                               & 32.8 & 30.8           & 70.8          & 38.5          & 37.5          & 85.9          & \textbf{52.7}  & \textbf{53.1}  & \textbf{90.6} \\
leggings                                                           & 18.2 & 20.5           & 49.0          & 30.8          & 37.3          & 70.7          & \textbf{36.9}  & \textbf{39.9}  & \textbf{73.7} \\
outerwear                                                          & 28.1 & 30.5           & 47.9          & 36.8          & 43.5          & 68.0          & \textbf{45.3}  & \textbf{51.9}  & \textbf{70.2} \\
pants                                                              & 28.2 & \textbf{33.33} & \textbf{51.5} & 23.9          & 27.3          & 42.4          & \textbf{30.4}  & \textbf{33.3}  & 48.5          \\
skirts                                                             & 62.3 & 68.0           & 80.2          & 64.5          & 71.2          & \textbf{86.5} & \textbf{73.3}  & \textbf{75.1}  & 85.5          \\
tops                                                               & 36.9 & 42.7           & 61.6          & 46.8          & 52.7          & 71.9          & \textbf{56.5}  & \textbf{57.9}  & \textbf{72.9} \\ \hline
\begin{tabular}[c]{@{}l@{}}Average over \\ categories\end{tabular} & 29.7 & 34.4           & 60.4          & 37.4          & 42.6          & \textbf{70.4} & \textbf{44.7}  & \textbf{47.4}  & 70.2          \\
\begin{tabular}[c]{@{}l@{}}Unconstrained \\ retrieval\end{tabular} & -    & -              & -             & 46.8          & 53.7          & \textbf{73.6} & \textbf{54.8*} & \textbf{57.1*} & 72.9*        
\end{tabular}
\end{center}
\end{table}

In Table \ref{tab:sota_comparison_df1} results on \DF dataset are presented.
Our model outperforms the results of \cite{dodds_learning_2018} in terms of Accuracy for all $k$ values. Especially Acc@1 noted largest relative boost.

\begin{table}[]
\begin{center}
\caption{Comparison of performance on \DF dataset using Acc@1, Acc@20 and Acc@50 metrics. Best performance for each metric is presented in bold}
\label{tab:sota_comparison_df1}
\begin{tabular}{ccc|ccc|ccc}
\multicolumn{3}{c|}{Current SOTA \cite{kuang_fashion_2019}} & \multicolumn{3}{c|}{Our Model} & \multicolumn{3}{c}{Our Model Re-ranking} \\
Acc@1    & Acc@20    & Acc@50    & Acc@1    & Acc@20    & Acc@50   & Acc@1    & Acc@20    & Acc@50        \\ \hline
25.73    & 64.38     & 75        & 37.8     & 77.2     & 84.1     & \textbf{40} & \textbf{79} & \textbf{85.5}
\end{tabular}
\end{center}
\end{table}
\begin{figure}
\begin{center}
\includegraphics[width=\textwidth]{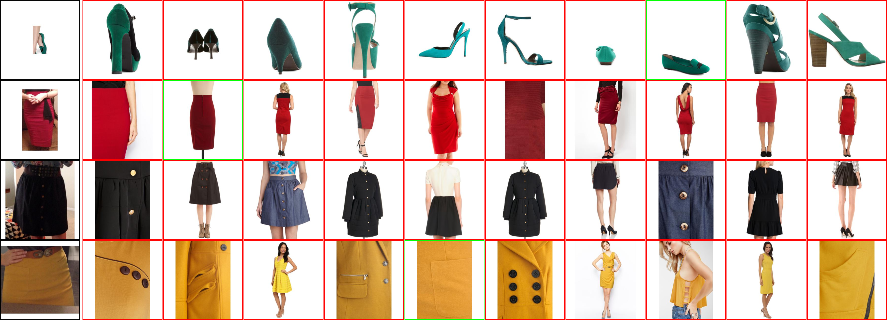}
\caption{Examples of retrieval on the \StoS dataset produced by our best model on 320x320 images. The images in the first column are query images, while the images on their right are the retrieval results with decreasing similarity towards the right side. Retrieval images with green border are the true match to the query. The top 10 most similar retrieval images are shown. It is worth noting that the retrieval was performed on the whole gallery dataset, with no pruning to the query item's category.}
\label{fig:s2s_retrieval_samples}
\end{center}
\end{figure}
\begin{figure}
\begin{center}
\includegraphics[width=\textwidth]{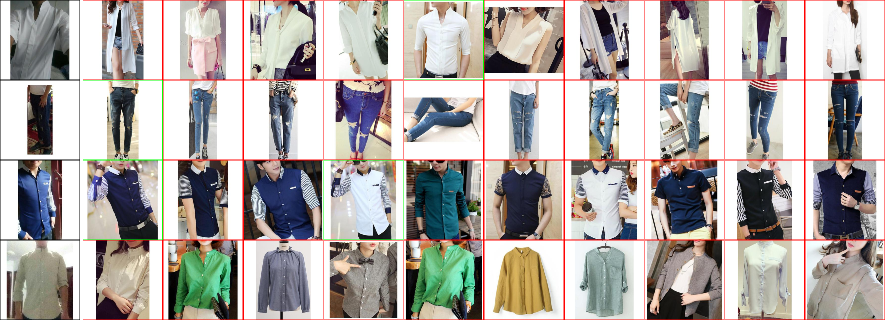}
\caption{Examples of retrieval on \DF dataset produced by our best model on 320x320 images. Retrieval and visualization settings are identical as in Figure \ref{fig:s2s_retrieval_samples}}.
\label{fig:df1_retrieval_samples}
\end{center}
\end{figure}

\subsection{Cross-domain Evaluation}

Inspired by the Person Re-Identification domain, where cross-domain evaluation is often performed to measure robustness of the models \cite{luo_strong_2019}, \cite{zhou_omni-scale_2019}, we decided to conduct such a test using our best performing RST model. 

Cross-domain evaluation consists of training a model on Dataset A and evaluation of its performance on Dataset B. In such settings training and test data distributions are different. 

In Table \ref{tab:cross_domain} we present the results of our cross-domain experiments, both with and without re-ranking. It can be seen that cross-domain performance of the RST model on both datasets is much lower than when model is trained and tested on the same dataset - same domain (See Table \ref{tab:sota_comparison_s2s} and Table \ref{tab:sota_comparison_df1}).  We suppose that such a large gap in performance between cross and same-domain evaluation is caused by a significant difference in data distribution of the two datasets. For example \DF contains only three categories: lower, upper and full-body, while \StoS contains 11 fine-grained clothing categories.

Even though cross-domain evaluation results are far from our best performing models, they are on a reasonable level when compared to the current state-of-the-art results for each dataset, thus, indicating that the RST model with \ti{Resnet50-IBN-A} backbone can learn meaningful representation of garments that can be transferred between domains to a large extent.


\begin{table}[]
\begin{center}
\caption{Comparison of the performance of the RST on cross-domain evaluation. DF $\,\to\,$ S2S means that the model was trained on \DF dataset and tested on the \StoS test set. For results marked with * see Section \ref{subsec:training}}
\label{tab:cross_domain}
\begin{tabular}{l|cccc|cccc}
              & \multicolumn{4}{c|}{DF 	$\,\to\,$ S2S} & \multicolumn{4}{c}{S2S $\,\to\,$ DF} \\
              & mAP   & Acc@1  & Acc@10  & Acc@20  & mAP   & Acc@1  & Acc@10  & Acc@20  \\ \hline
No re-ranking & 26.2  & 37.7   & 49.5    & 53.3    & 20.9  & 18.6   & 40.7    & 47.2    \\
Re-ranking    & 27.9*  & 37.4*   & 48.8*    & 51.7*    & 23.5  & 20.5   & 43.7    & 49.6   
\end{tabular}
\end{center}
\end{table}

\section{Conclusions}\label{sec:conclusions}

In this paper we examined the analogies and differences between Person Re-Identification research field and Fashion Retrieval, then examined the transferability of ReID models, and performed adjustments necessary to make them work for fashion items.

Using the approach proposed by \cite{luo_strong_2019} using only global features, a number of improvements and learning optimizations we achieved 54.8 mAP, 72.9 Acc@20 on \StoS and 47.3 Acc@1, 79.0 Acc@20 on \DF dataset, establishing new state-of-the-art results for both. The performance on \StoS seems particularly robust, compared to previous state-of-the-art as our results were achieved on images several times smaller than the previous top-scoring models.

By analogy to \cite{luo_strong_2019} we consider the results achieved by our model as a strong baseline for further clothes retrieval research. Despite a simple architecture, the baseline model outperforms previous solutions dedicated and customized specifically for fashion retrieval systems. Our results are a clear indication that field of fashion retrieval can benefit from research established for Person Re-Identification.
Additionally, we performed supplementary experiments and showed that quadruplet loss may bring further improvements at a negligible cost, and that re-ranking can further boost performance.
Finally, we introduced cross-domain evaluation into clothes retrieval research to test robustness of fashion retrieval models. 


\clearpage

%
%
\bibliographystyle{splncs04}
\bibliography{my_bib}

\pagestyle{headings}
\mainmatter
\def\ECCVSubNumber{xxx}  

\title{Supplementary material for: A Strong Baseline for Fashion Retrieval with Person Re-Identification models} 



\titlerunning{A Strong Baseline for Fashion Retrieval with Person Re-Identification models}
%
\author{Mikolaj Wieczorek\inst{1} \and
Andrzej Michalowski\inst{1} \and
Anna Wroblewska\inst{1,2}\orcidID{0000-0002-3407-7570} \and 
Jacek Dabrowski\inst{1}\orcidID{0000-0002-1581-2365}}
\authorrunning{M. Wieczorek et al.}
%
\institute{Synerise \and 
Faculty of Mathematics and Information Science, Warsaw University of Technology}


\maketitle

In the following sections of supplementary material we provide loss function definitions used in fashion retrieval domain (Section~\ref{sec:supp:loss_functions}) along with metrics used (Section~\ref{sec:supp:metrics}). Then we describe two open datasets used in our tests which are commonly used in the domain (Section~\ref{sec:sup:datasets}). We also demonstrate a few problems with the datasets along with our adjustments for the community to use them in more convenient way using the popular COCO format.

Finally we list a few examples of outputs of our models, and also examples with re-ranking technique and without them (Section~\ref{sec:supp:outputs}).

In the following sections of supplementary material we provide loss function definitions used in fashion retrieval domain (Section~\ref{sec:supp:loss_functions}) along with metrics used (Section~\ref{sec:supp:metrics}). Then we describe two open datasets used in our tests which are commonly used in the domain (Section~\ref{sec:sup:datasets}). We also demonstrate a few problems with the datasets along with our adjustments for the community to use them in more convenient way using the popular COCO format.

Finally we list a few examples of outputs of our models, and also examples with re-ranking technique and without them (Section~\ref{sec:supp:outputs}).

\section{Loss functions}
\label{sec:supp:loss_functions}


In the image retrieval task there are two loss functions commonly used: classification and triplet loss. Therefore, prevailing number of works in the image retrieval domain use a combination of a classification and a triplet loss for training deep learning models. Classification loss function is used to identify exact id of a person/garment (i.e. images of the same person/garment have the same id). It is a standard loss in classification tasks and in our case it is cross-entropy loss. Also all of the considered models use either of these two loss functions. 

Deep metric learning treats the image retrieval problem somewhat as a clustering or ranking problem and optimizes the mutual arrangement of the embeddings in space. One of the most widely used approaches that pulls the same class closely and pushes away other class embeddings is using a triplet loss \cite{schroff_facenet:_2015} for training neural network. 

Triplet loss is formulated as follows:
\begin{equation}
    \mathcal{L}_{triplet} = \left[ \left\Vert f(A)-f(P) \right\Vert_2^2 - \left\Vert f(A)-f(N) \right\Vert_2^2 + \alpha  \right]_+\label{eq_1}
\end{equation}
where $[z]_+ = max(z,0)$ and $f$ denotes learnt embedding function applied to all data points. 

A triplet loss consists of an anchor image (a query) $A$, a positive example $P$ -- the other image of the same object (in this paper -- clothing item) present in the $A$ image -- and negative sample $N$, which is an image of a different object from that shown in the image $A$. 

Learning NN with triplet loss minimizes the intra-class distance, between anchor and positive samples, and maximizes inter-class distance, between anchor and negative samples. The triplet loss proved to achieve state-of-the-art performance and became a standard in similarity learning tasks \cite{wang_learning_2014}, \cite{schroff_facenet:_2015}, \cite{hermans_defense_2017}.

In triplet loss strategy, the method of creating triplets is an important part of the training and influences the model performance immensely \cite{wu_sampling_2018}. In \cite{schroff_facenet:_2015} authors used semi-hard triplets, where negative samples are further away from the anchor than the positive samples, but still the loss value is positive, thus, it allows learning.

Most of the works we examined use online hard negative sampling to form a training triplet. This methods select such data points, so that the negative sample is closer to the anchor than the positive sample. As a results, the neutral network is given only the triplets that maximize the value of the loss function, therefore it is called 'hard'. This method of creating triplets proved to perform better than other sampling methods and is used in numerous works \cite{shrivastava_training_2016}, \cite{zhang_joint_2016}, \cite{luo_strong_2019}, \cite{vishvakarma_mildnet:_2019}.

To further improve the triplet loss some authors either extends the number of tuples in the loss \cite{chen_beyond_2017}, \cite{sohn_improved_2016}, \cite{wang_ranked_2019} or/and propose novel sampling methods \cite{harwood_smart_2017}, \cite{wu_sampling_2018}.
However, the reported improvements are not high, thus, we did not use them in our experiments.


Triplet, and in general n-tuple-loss, aims to properly arrange embeddings in an n-dimensional space. While the triplet loss is a common choice in retrieval/ranking tasks, we also examined the quadruplet loss and its influence on the performance. Our implementation of the quadruplet loss follows one found in \cite{chen_beyond_2017}:
\begin{equation}
\begin{split}
    \mathcal{L}_{quad} = \left[ \left\Vert f(A)-f(P) \right\Vert_2^2 - \left\Vert f(A)-f(N_1) \right\Vert_2^2 + \alpha_1 \right]_+ + \\
                         \left[ \left\Vert f(A)-f(P) \right\Vert_2^2 - \left\Vert f(N_2)-f(N_1) \right\Vert_2^2 + \alpha_2 \right]_+\label{eq_2}
\end{split}
\end{equation}
where the first term is the same as in Equation \ref{eq_1}, thus, it takes care of the pull-push relation between anchor, positive and negative samples. The second term demands the intra-class distance to be smaller than the maximum inter-class distance in respect to a different probe - $N_2$ ($N_2 \ne N_1 \Rightarrow$ $N_2$ and $N_1$ represents different garments/IDs). $\alpha_1$ and $\alpha_2$ are the margin values, which are set dynamically as in \cite{chen_beyond_2017}.

In \cite{luo_strong_2019} additionally center loss \cite{leibe_discriminative_2016} is used as one of the training tricks. It aims to pull same class embeddings together as the n-tuple-loss considers only relative distance between embeddings neglecting the distance absolute values. Center loss alleviate this problem by penalizing the distance between embeddings and their id/class center.
Formula for center loss is as follows:
\begin{equation}
     \mathcal{L}_{center} = \frac{1}{2} \sum_{j=1}^{B} \left\Vert \boldsymbol{f}_{t_j} - \boldsymbol{c}_{y_j}\right\Vert_2^2
\end{equation}
where $y_j$ denotes label of $j$-th image in the mini-batch. $B$ is the batch size, $\boldsymbol{f}_{t_j}$ is an embedding of $j$-th image and $\boldsymbol{c}_{y_j}$ is the center of $y_j$-th class features center.

\section{Metrics in fashion retrieval}
\label{sec:supp:metrics}

To evaluate the performance of our approach we used metrics that we found most often in the related papers. Most widely used metric in retrieval tasks is $Accuracy@k$ ($Acc@k$), formulated as:
\begin{equation}
     Acc@k = \frac{1}{N} \sum_{i=1}^N 1 \left[ S^+_q \cap S^K_q \right]\label{metric:acc_at_k},
\end{equation}
where $N$ is the number of queries and $1 \left[ S^+_q \cap S^K_q \right]$ is an indicator function, which evaluates to 1 if the ground-truth image is within top-k retrieved results.
$k$ is usually set from 1 to 20. The metric measures if the retrieved item was among top-k proposals.

Second metric that we encountered in the papers was $mAP$, which is a mean average precision, that shows how well the retrieval is done on average. Though $mAP$ values were rarely reported in clothes retrieval papers we decided to use this metric in our experiments along $Acc@k$. 

\section{Datasets}
\label{sec:sup:datasets}

In this section we describe datasets used for evaluation. Apart from describing their statistics, we also explain the process of reformatting them and how they were processed during our experiments.

\subsection{\StoS}
The dataset was introduced by \cite{kiapour_where_2015} and became one of the most widely used datasets for evaluating clothes retrieval solutions. Therefore there is an abundance of works that present their results on the dataset, thus, providing a strong benchmark for our methods. It contains 404,683 shop photos and 20,357 street photos depicting 204,795 distinct clothing items \cite{kiapour_where_2015}. To allow compatibility across datasets and models we tested, we transformed the \StoS dataset to COCO-format, while keeping original train/test split and categorization. Annotations in COCO-format are available on our GitHub\footnote{\label{footnote:our_github}\url{Link will be released in the final submission}}.

In contrast to some authors \cite{kuang_fashion_2019} we decided not to perform any data cleaning or hand-curating images/annotations, even though we encountered some erroneous annotations such as multiple annotations for a single-item image or bounding boxes placed in 'random' places (see examples in Fig.~)).

We made such decision to allow a fair comparison with \cite{kucer_detect-then-retrieve_2019}, which we found to have best performance on \StoS dataset, while it does not mention any data cleaning.

\subsection{\DF}
The dataset contributed by \cite{liu_deepfashion:_2016} contains over 800,00 images, but for our task we only used \textit{Consumer-to-shop Clothes Retrieval} subset that consists of 33,881 distinct clothing items and total of 239,557 images, creating 195,540 pairs. We used results found in \cite{dodds_learning_2018} as our benchmark, since their were the best we found. 

\begin{figure}
 \begin{center}
 \includegraphics[width=\textwidth]{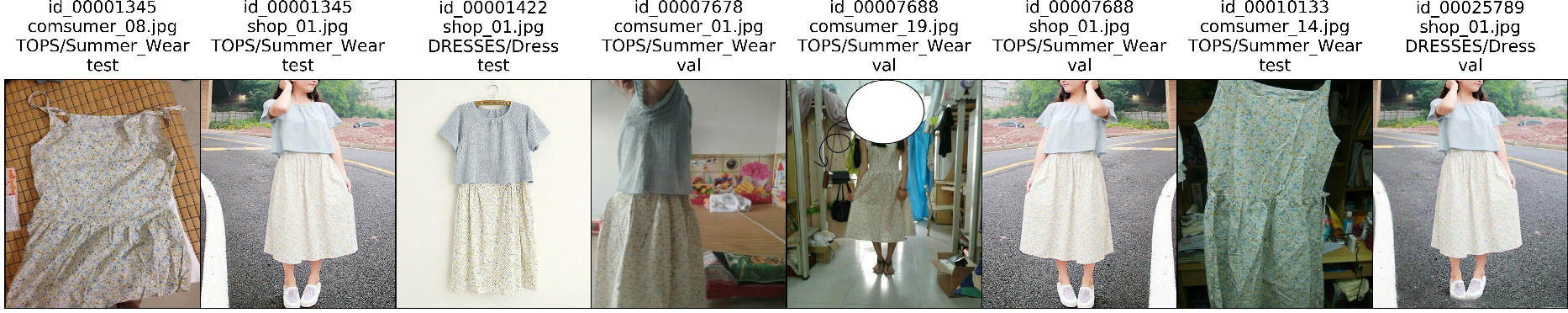}
 \caption{A collection of photos that depict, we believe, the same clothing item - dress visible in the left most photo. Above each photo there are four pieces of information; from the top: item id, file name, category, subset name. Despite the fact that item ids should be unique for distinct garments, it seems that the same item have various ids assigned, which results in erroneous retrieval results presented in Figure \ref{fig:df1_duplicates_visrank_1}}
 \label{fig:df1_duplicates_sample_1}
 \end{center}
 \end{figure}

\begin{figure}
 \begin{center}
 \includegraphics[width=\textwidth]{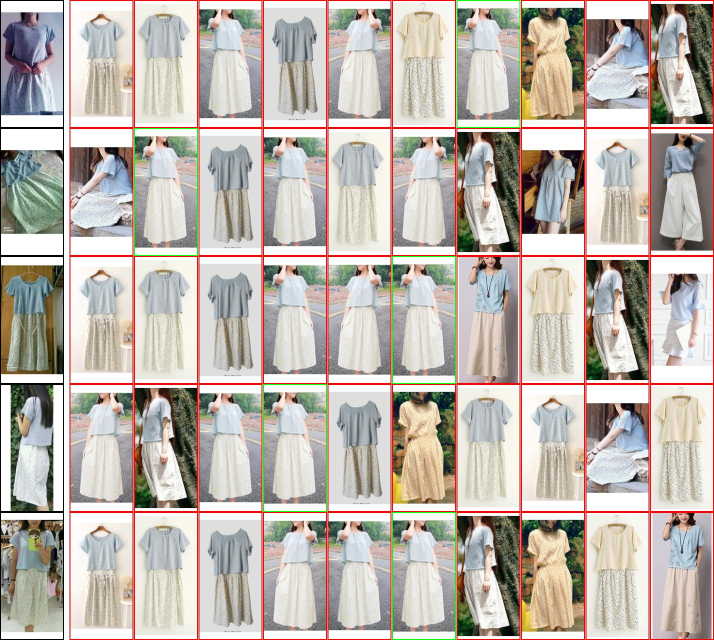}
 \caption{Examples of retrieval for the query images with product ids from presented in Figure \ref{fig:df1_duplicates_sample_1} produced by our best model on 320x320 images. The images in the first column are query images, while the images on their right are the retrieval results with decreasing similarity towards the right side. Retrieval images with green border are the true match to the query. The top 10 most similar retrieval images are shown. It can be seen that some images that are just mirrored copies of the same image, yet only one of them is deemed as a true match. We believe it is an error in data annotation, which understates real retrieval performance.}
 \label{fig:df1_duplicates_visrank_1}
 \end{center}
 \end{figure}

\begin{figure}
 \begin{center}
 \includegraphics[width=0.7\textwidth]{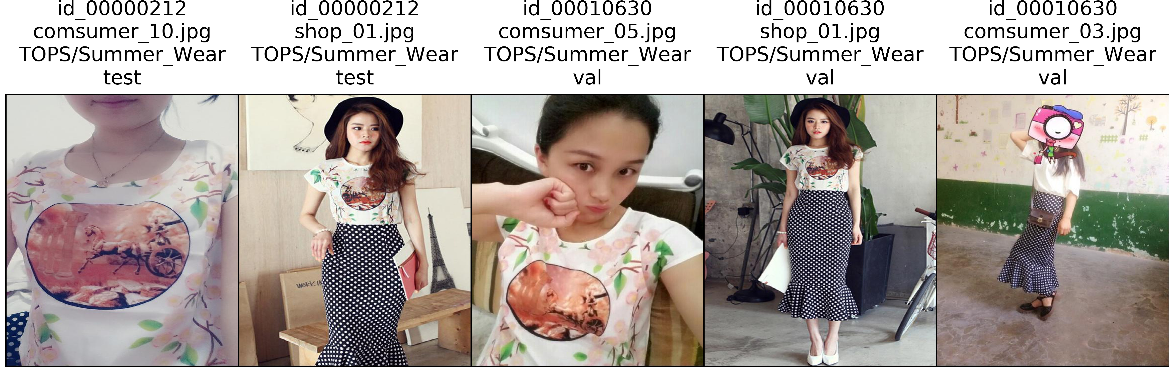}
 \caption{A collection of photos that depict, we believe, the same clothing item - top visible in the left most photo. Above each photo there are four pieces of information; from the top: item id, file name, category, subset name. Despite the fact that item ids should be unique for distinct garments, it seems that the same item have various ids assigned, which results in erroneous retrieval results presented in Figure \ref{fig:df1_duplicates_visrank_2}. Interestingly, the right most photos depicts a top that is plain white, which also seems to be incorrect compared the rest.}
 \label{fig:df1_duplicates_sample_2}
 \end{center}
 \end{figure}

\begin{figure}
 \begin{center}
 \includegraphics[width=\textwidth]{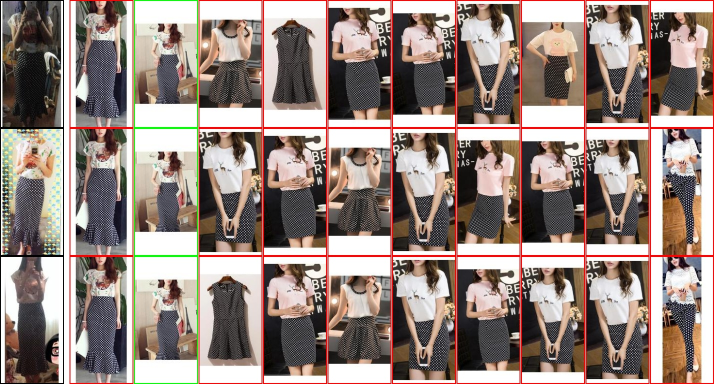}
 \caption{Examples of retrieval for the query images with product ids from presented in Figure \ref{fig:df1_duplicates_sample_2} produced by our best model on 320x320 images. The images in the first column are query images, while the images on their right are the retrieval results with decreasing similarity towards the right side. Retrieval images with green border are the true match to the query. The top 10 most similar retrieval images are shown. It can be seen that all results have first and second image the same, while in all cases only the latter is correct even though the former is also from 'top' category, thus, it seems to be pertaining exactly the same garment.}
 \label{fig:df1_duplicates_visrank_2}
 \end{center}
 \end{figure}

Similarly to \StoS dataset, \DF is also not free from some defects, which we show in Figures \ref{fig:df1_duplicates_sample_1}, \ref{fig:df1_duplicates_visrank_1}, \ref{fig:df1_duplicates_sample_2}, \ref{fig:df1_duplicates_visrank_2}.

In \textit{Consumer-to-Shop} subset of \DF, we found out that the same products and even the same images were assigned different product identifiers. As a result, the retrieval performance is falsely understated compared to the real performance. Two examples of such errors are presented in the supplementary materials to this paper

\section{Code improvements}\label{sec:code_improvements}

As we mentioned in the main text, we encountered problems with vRAM and RAM overflow caused by the size of the datasets we tested. 
The RST model contains fully connected layers used for classification of identities/clothes. While ReID datasets the RST model was tested on – \textit{Market-1501}, \textit{DukeMTMC-reid} contain 1501 and 1812 unique identities, the fashion datasets have an order of magnitude more identities (clothes) – roughly 10000-15000. As a result, the FC layer needs thousands neurons instead of hundreds, thus, it requires much more video RAM (vRAM. To address this problem we introduced two independent solutions:
\begin{enumerate}
    \item Gradient accumulation - it allows to use smaller mini-batches in constrained vRAM settings when the mini-batch either do not fit into GPU memory or is too small causing the gradient descent to be volatile and prevent model from converging.  Gradient accumulation splits original mini-batch into sub-mini-batches, feeds them into network, compute gradients, but the model weights are updated after all sub-mini-batches. 
    \item ModelParallel mode - it is a distributed training technique that splits a single model across different devices. The data and gradients are moved between devices during forward and backward pass. It was implemented to allow us to test larger backbones that would not fit into a single GPU, use larger mini-batches and, thus speed up the training.
\end{enumerate}

In the code version we used, Resnet-50-IBNs were not yet implemented, therefore we implemented both A and B variants by ourselves based on original implementation. Additionally, we expanded available backbones with the whole EfficientNet family based on the implementations available here.\footnote{\url{https://github.com/lukemelas/EfficientNet-PyTorch}}

During evaluation step, especially when performing re-ranking, we encountered problem with RAM consumption. Again, the problem arises from the size of the clothes retrieval datasets we tested. 
To tackle the problem we introduced three solutions:
\begin{enumerate}
    \item We introduced batch processing during both creating images' embeddings, to avoid vRAM overflow, and during computation of distance matrix for tens of thousands images. Originally, both operations were performed in one go.
    \item Conducting evaluation with re-ranking for single categories for \StoS \\ was still problematic due to large RAM requirements, so we used batch processing again, but, we appended intermediate results from batch computations of distance matrix to a file in a hard drive. During re-ranking itself we used Numpy function \textit{memmap} to avoid reading the whole matrix into RAM, while still allowing RAM-like processing.
    \item Unfortunately, evaluation with re-ranking and without category constraint was still impossible for \StoS, as the whole distance matrix – over 400,000x400,000 floats, was again too big even when using \text{memap} function. 
    We decided to conservatively estimate the results by calculating weighted average over categories and deducting a penalty term. The penalty term was computed for each metric separately using results from the model variant without re-ranking, as the maximum difference, between the weighted average over categories and \textit{Unconstrained Retrieval} values.
\end{enumerate}


Finally, we added \textit{Accuracy@k} computation for specified \textit{k} and a script that at the end of evaluation creates visualization of the results. It plots query image and top-k retrieved images.






\section{Our result examples}
\label{sec:supp:outputs}


\begin{figure}
\begin{center}
 \includegraphics[width=\textwidth]{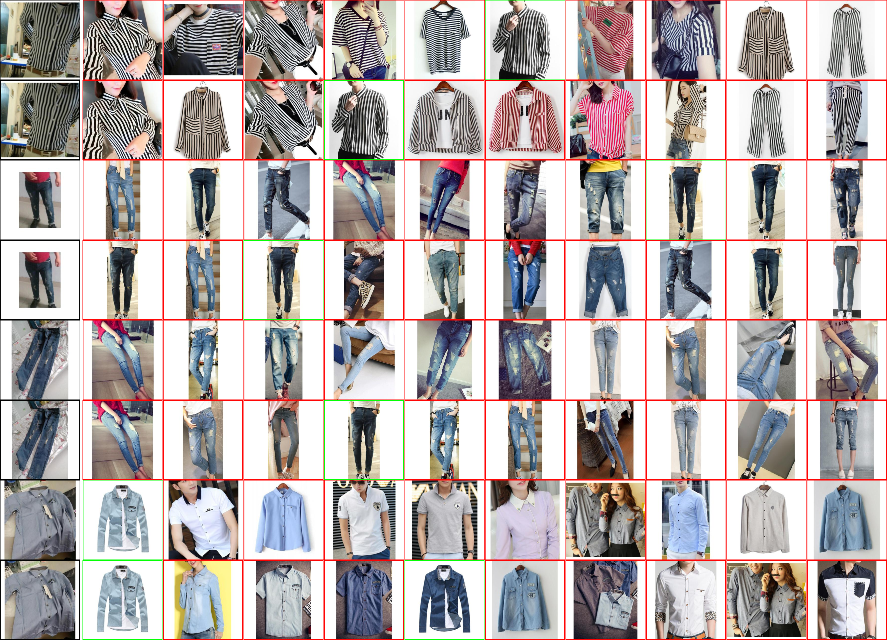}
 \caption{Examples of retrieval on \DF dataset produced by our best model on 320x320 images. The images in the first column are query images, while the images on their right are the retrieval results with decreasing similarity towards the right side. Retrieval images with green border are the true match to the query. The top 10 most similar retrieval images are shown. Two retrieval results are shown for each query image, one without and one with re-ranking. The top result from a pair is without re-ranking.}
 \label{fig:df1_rerank_0}
 \end{center}
 \end{figure}

 \begin{figure}
 \begin{center}
 \includegraphics[width=\textwidth]{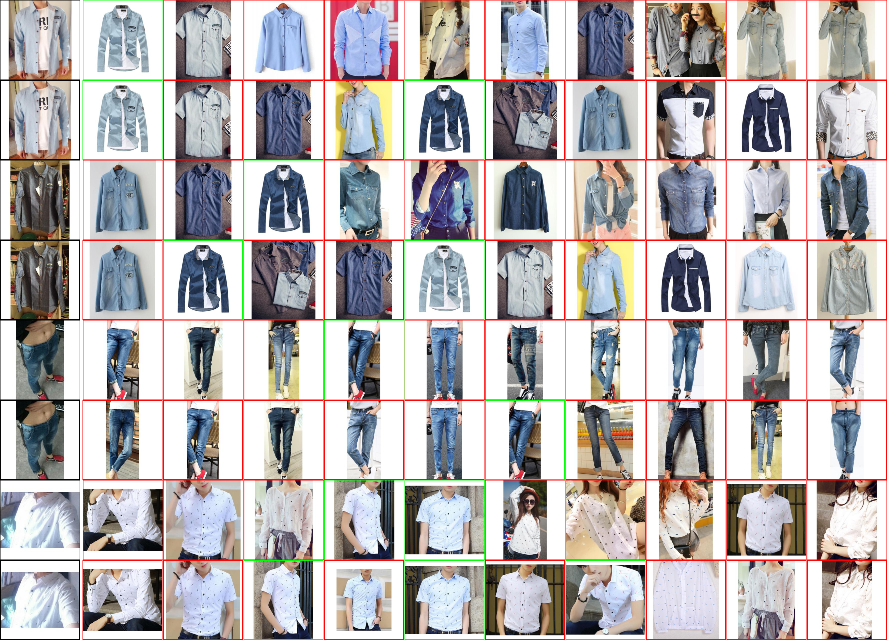}
 \caption{More retrieval results on \DF dataset without and with re-ranking.}
 \label{fig:df1_rerank_1}
 \end{center}
 \end{figure}

 \begin{figure}
 \begin{center}
 \includegraphics[width=\textwidth]{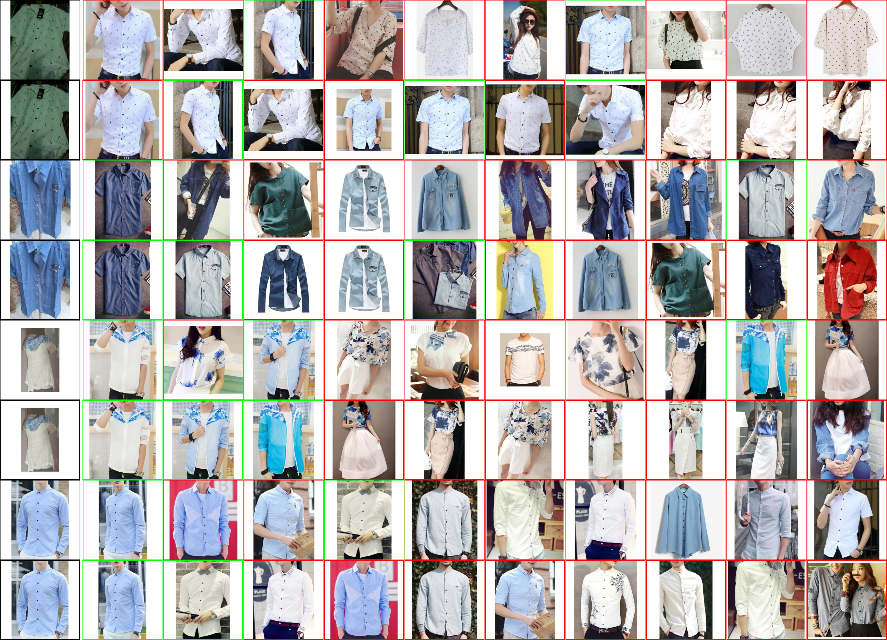}
 \caption{More retrieval results on \DF dataset without and with re-ranking.}
 \label{fig:df1_rerank_2}
 \end{center}
 \end{figure}


\begin{figure}
 \begin{center}
 \includegraphics[width=\textwidth]{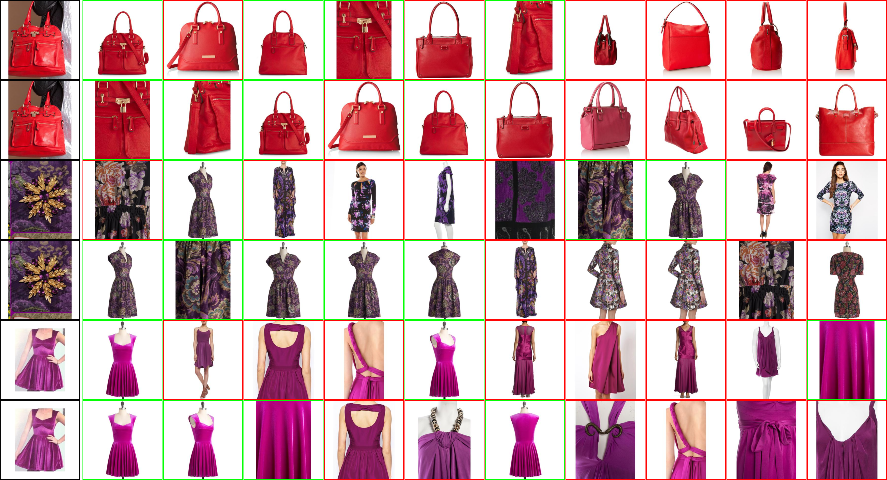}
 \caption{Examples of retrieval on \StoS dataset produced by our best model on 320x320 images. The images in the first column are query images, while the images on their right are the retrieval results with decreasing similarity towards the right side. Retrieval images with green border are the true match to the query. The top 10 most similar retrieval images are shown. Two retrieval results are shown for each query image, one without and one with re-ranking. The top result from a pair is without re-ranking.}
 \label{fig:s2s_rerank_0}
 \end{center}
 \end{figure}

\begin{figure}
 \begin{center}
 \includegraphics[width=\textwidth]{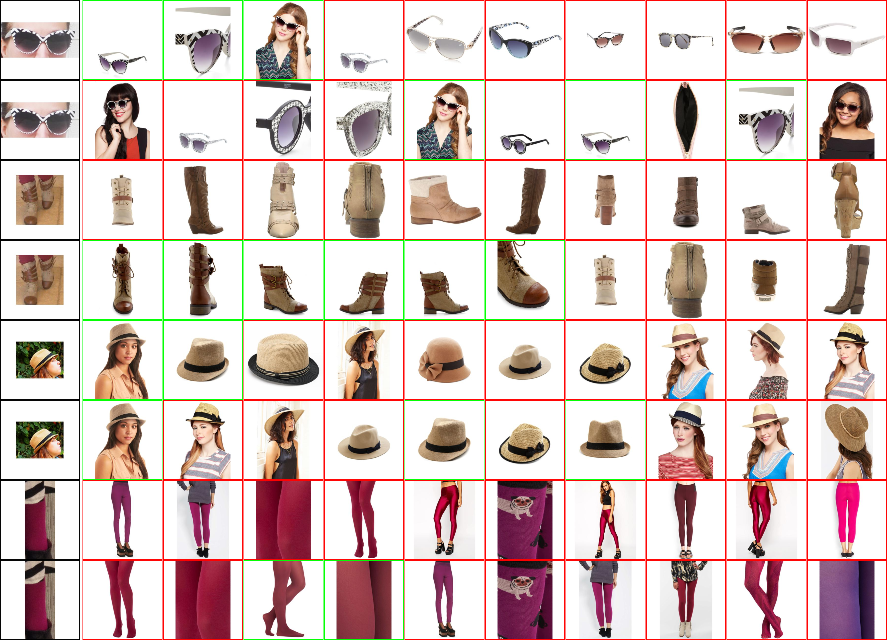}
 \caption{More retrieval results on \StoS dataset without and with re-ranking.}
 \label{fig:s2s_rerank_1}
 \end{center}
 \end{figure}

 \begin{figure}
 \begin{center}
 \includegraphics[width=\textwidth]{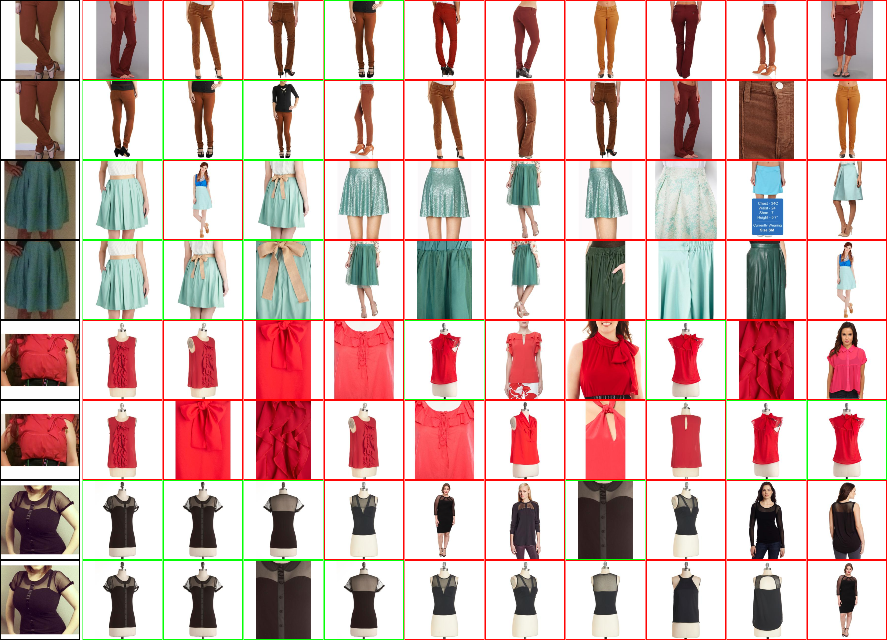}
 \caption{More retrieval results on \StoS dataset without and with re-ranking.}
 \label{fig:s2s_rerank_2}
 \end{center}
 \end{figure}

\end{document}